\documentclass{article}
\usepackage{ijcai22}

\usepackage{times}
\usepackage{soul}
\usepackage{url}
\usepackage[hidelinks]{hyperref}
\usepackage[utf8]{inputenc}
\usepackage[small]{caption}
\usepackage{graphicx}
\usepackage{amsmath}
\usepackage{amsthm}
\usepackage{booktabs}
\usepackage{algorithm}
\usepackage{algorithmic}
\usepackage{amssymb}        

\usepackage{multirow}
\usepackage{booktabs}
\usepackage{tabularx}
\usepackage{makecell}
\usepackage{threeparttable}

\title{Diversity is Strength: Mastering Football Full Game with Interactive Reinforcement Learning of Multiple AIs}

\author{
	Chenglu Sun\and
	Shuo Shen\and
	Sijia Xu\And
	Weidong Zhang
	\affiliations
	AI Lab, Netease\\
	\emails
	\{sunchenglu, shenshuo01, xusijia, zhangweidong02\}@corp.netease.com
}

\begin{document}

\maketitle

\begin{abstract}
Training AI with strong and rich strategies in multi-agent environments remains an important research topic in Deep Reinforcement Learning (DRL). The AI's strength is closely related to its diversity of strategies, and this relationship can guide us to train AI with both strong and rich strategies. To prove this point, we propose Diversity is Strength (DIS), a novel DRL training framework that can simultaneously train multiple kinds of AIs. These AIs are linked through an interconnected history model pool structure, which enhances their capabilities and strategy diversities. We also design a model evaluation and screening scheme to select the best models to enrich the model pool and obtain the final AI. The proposed training method provides diverse, generalizable, and strong AI strategies without using human data. We tested our method in an AI competition based on Google Research Football (GRF) and won the 5v5 and 11v11 tracks. The method enables a GRF AI to have a high level on both 5v5 and 11v11 tracks for the first time, which are under complex multi-agent environments. The behavior analysis shows that the trained AI has rich strategies, and the ablation experiments proved that the designed modules benefit the training process.
\end{abstract}

\section{Introduction}
\label{Section1}
Enhancing the strength of Artificial Intelligence (AI) is a prominent concern in Deep Reinforcement Learning (DRL) research. Self-play, a widely adopted training paradigm \cite{silver2016mastering,silver2018general}, has shown promise in improving DRL-based AI strength. However, determining whether the agent is making steady progress with self-play training can be challenging due to a common DRL phenomenon known as ``non-transitivity'' \cite{balduzzi2019open}. Non-transitivity arises from the fact that during self-play training, the opponent models are usually the latest versions of the agent itself. As a result, the agent tends to focus only on discovering and exploiting new strategies that can beat its current strategy. However, during this exploration, the previous strategy may be forgotten, leading to insufficient model generalization and increased vulnerability to exploitation by other agents. Another possible reason for non-transitivity is the limitation of state distribution during self-play training, which may cause the model to be unable to learn strategies in other states or perform well in states with disturbance \cite{gleave2019adversarial}. To address this problem, recent studies have employed the history model pool (HMP) \cite{berner2019dota,li2020suphx,guss2021minerl,ye2020mastering} to mitigate the phenomenon. The HMP approach involves maintaining a pool of previous versions of the model, allowing the agent to train against a more diverse set of opponents. While HMP can improve the performance of trained models, it does not guarantee exceptional quality since the models in the pool are generated naturally over time and may lack diversity and robustness in their strategies. Thus, the diversity of strategies of the training models cannot be guaranteed either.

AI with limited strategies is vulnerable to quick defeat by identifying its weaknesses, and the lack of diversity in strategies may lead to another issue: the AI may become less applicable. Applicability has gradually become a crucial criterion for AIs \cite{risi2020chess}, and strong AI in a single scenario has failed to meet the ever-increasing demand for AI in various application scenarios, such as game AI with different styles and characteristics \cite{yannakakis2018artificial}. The primary reason for the lack of diversity is that most DRL methods only adhere to the reward maximization principle \cite{tian2019elf} and tend to overfit the states in a limited environment, resulting in inflexible behaviors. In DRL, a common technique to enhance strategy diversity is to create multiple sets of reward shaping and apply them to model training \cite{de2021configurable,mysore2023multi}. However, the strength of these strategies cannot be guaranteed.

Therefore, enhancing strategy diversity while maintaining strength is crucial in training AI through DRL. To this end, we propose a novel DRL training framework called Diversity is Strength (DIS) that enables the AI to acquire rich strategies, attain better generalization, and reach a higher level. DIS improves the strength and strategy diversity of the training model by fostering diversity among opponents. The framework simultaneously trains three types of AI: the main agent, the policy explorer, and the method explorer. The main agent is the core AI for long-term training and maintenance, the policy explorer explores different training settings derived from the main agent, and the method explorer explores completely different DRL methods. The HMPs of these three AIs are enriched by adding the short-term and long-term history models. Moreover, a model evaluation and screening scheme is designed to periodically select the best models to be added to the HMPs. Through these designs, the HMPs of all three AIs can be shared, forming a complex and interconnected HMP structure that ensures diversity among opponents.

\begin{figure}[h]
	\centering
	\includegraphics[scale=0.098]{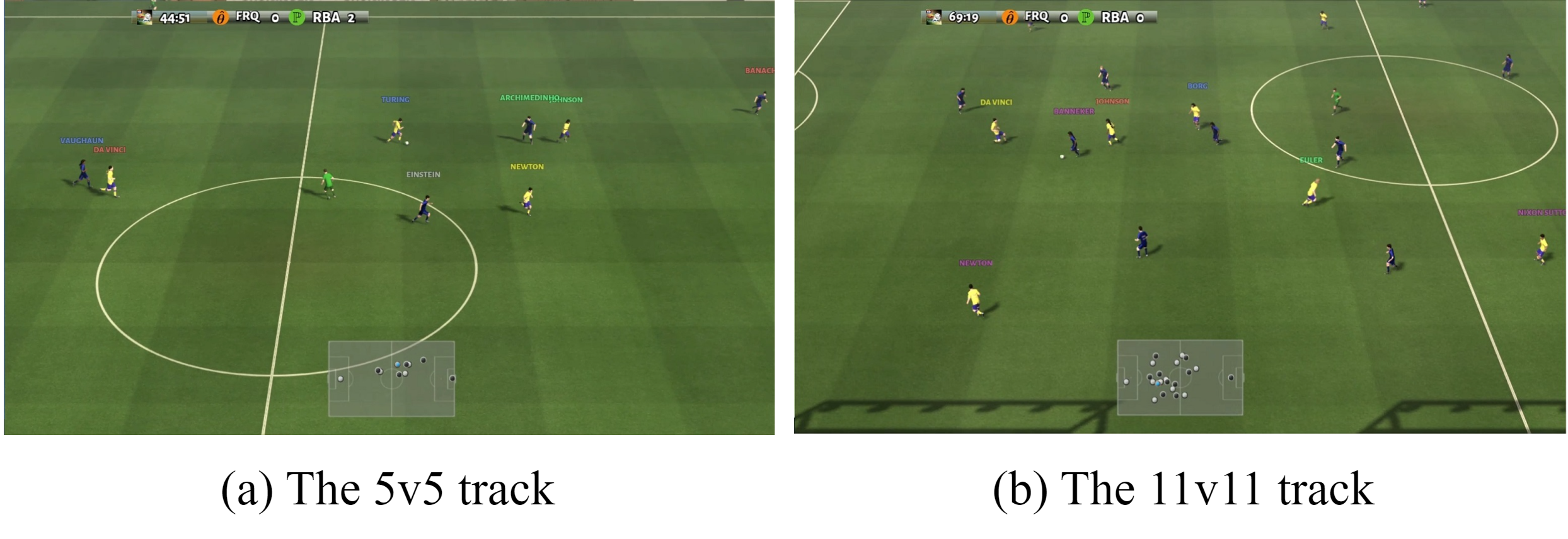}
	\caption{The 5v5 track and 11v11 track in the GRF environment. The AI simultaneously controls all players in the full game, while only the keeper in 5v5 track is controlled by the game environment.}
	\label{fig:1}
\end{figure}

The AI trained by DIS has won the championships in all tracks of the Institute of Electrical and Electronics Engineers Conference of Game (IEEE CoG) 2022 Football AI Competition. This competition was held on the well-established AI testing platform, Google Research Football (GRF) \cite{kurach2020google}, and was participated by numerous researchers. The competition included two tracks, 5v5 and 11v11, both of which are multi-agent battle environments (refer to Figure \ref{fig:1}). The victories in both tracks demonstrate the efficacy of our proposed method. A series of experiments have shown that the method can facilitate a steady improvement and a rich set of strategies in the trained AI. The main contributions of this study are as follows: (1) Proposing a novel DRL training framework to train AIs in a complex multi-agent environment, which can enable the AIs to develop strong and rich strategies; (2) Conducting sufficient ablation and comparative experiments to demonstrate the effectiveness of each component of the proposed method.

\section{Related Works}
\label{Section:2}
Many successful DRL-based models have been trained using the self-play method with HMP. For instance, OpenAI Five \cite{berner2019dota} plays against historical models with a 20\% probability, and the history models are selected based on their win rates with respect to the training model. The authors stated that using history models could prevent strategy collapse, and eventually, they achieved a Dota2 AI that defeated the world champion team. In AlphaStar \cite{vinyals2019grandmaster}, a DRL training framework called League Training (LT) was proposed to improve the AI level. Three agents, trained using the same method, but playing against different training opponents, were employed in LT. By increasing the complexity of the opponents, AlphaStar achieved top-level performance against human players. However, the periodic resetting of two agents to a supervised learning (SL) based agent limits the intensity and diversity of AI strategies. Furthermore, the overlap between the opponents of the three agents reduces training efficiency. Many players have noted that although the two highly competent AIs have achieved human-level proficiency, they lack variation in their response to different scenarios. This makes their weaknesses easily exploitable by human players. For example, OpenAI Five relies heavily on a fixed strategy that differs significantly from human players, making it easily identifiable as an AI agent. To counter this issue, AlphaStar employed an SL-based model trained with expert data as initial model to emulate human player strategies. Additionally, Kullback-Leibler (KL) Divergence \cite{joyce2011kullback} was introduced in the subsequent RL training process to further refine the approximation. Despite these efforts, the strategy of AlphaStar remains close to that of the SL-based model, limiting the variety of trained strategies. This lack of diversity can hinder the development of AI capabilities and limit its applicability to a wider range of scenarios. To address this, we have developed an innovative DRL training framework that employs an elaborate structure of HMPs. In this framework, opponents are generated not only from the AI's history model during different training periods but also from other AIs with differing strategies. The aim is to equip the AI with a broad range of powerful and diverse strategies.

\section{Method}
\label{Section3}
In this section, we first introduce the proposed DIS training framework. We then detail how we train our AIs, and finally present the model evaluation process.

\subsection{DIS Architecture}
\label{Section3.1}
\begin{figure}[h]
	\centering
	\includegraphics[scale=0.16]{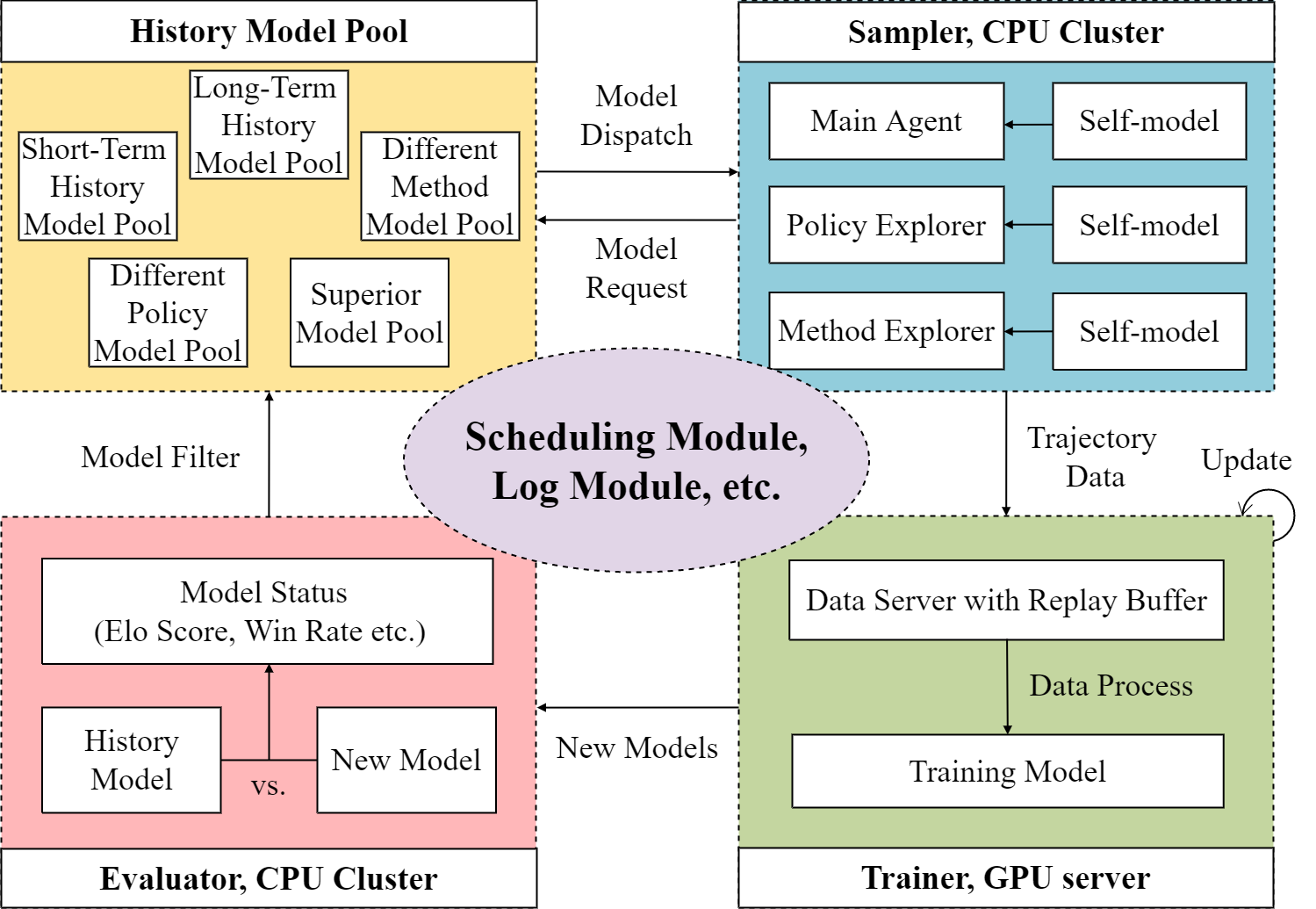}
	\caption{Overview of the DIS training framework.}
	\label{fig:2}
\end{figure}

The DIS framework consists of four main components: the sampler, trainer, HMP, and evaluator, as illustrated in Figure \ref{fig:2}. The sampler involves three types of designed AIs, namely the main agent, policy explorer, and method explorer. The main agent and policy explorer employ the same training algorithm, while the training configurations of the main agent remain constant throughout the training process. And the policy explorer generates models with different strategies by utilizing various training settings, including designs of reward shaping, hyperparameters, input features, and training scenarios. The method explorer obtains models with different strategies by training with distinct algorithms. The diverse training settings and algorithms are the primary reasons for producing varied model effects. Consequently, these AIs can generate trajectories with different state distributions, which are sent to the data server as a replay buffer. The trainer updates the model with the processed data extracted from the data server. During the training, the evaluator enables each new model to play against the history models to validate its performance, using metrics such as the Elo score \cite{albers2001elo} and win rate. The new models are then transmitted to the elaborated HMPs which will be introduced in next paragraph. The AIs on the samplers can choose the current self-model or history models from the HMPs as opponents, according to our designed model sampling method, which will be detailed in Section 3.2.3. The scheduling module of the DIS can automatically allocate resources for the training process based on the configuration. Through this cyclic process, the training architecture can generate abundant training data and diverse models, which continuously improve the AIs, as demonstrated in Section 4.1. The trainer and HMP are built on GPU servers, while the sampler and evaluator are built using CPU clusters.

\begin{figure}[h]
	\centering
	\includegraphics[scale=0.15]{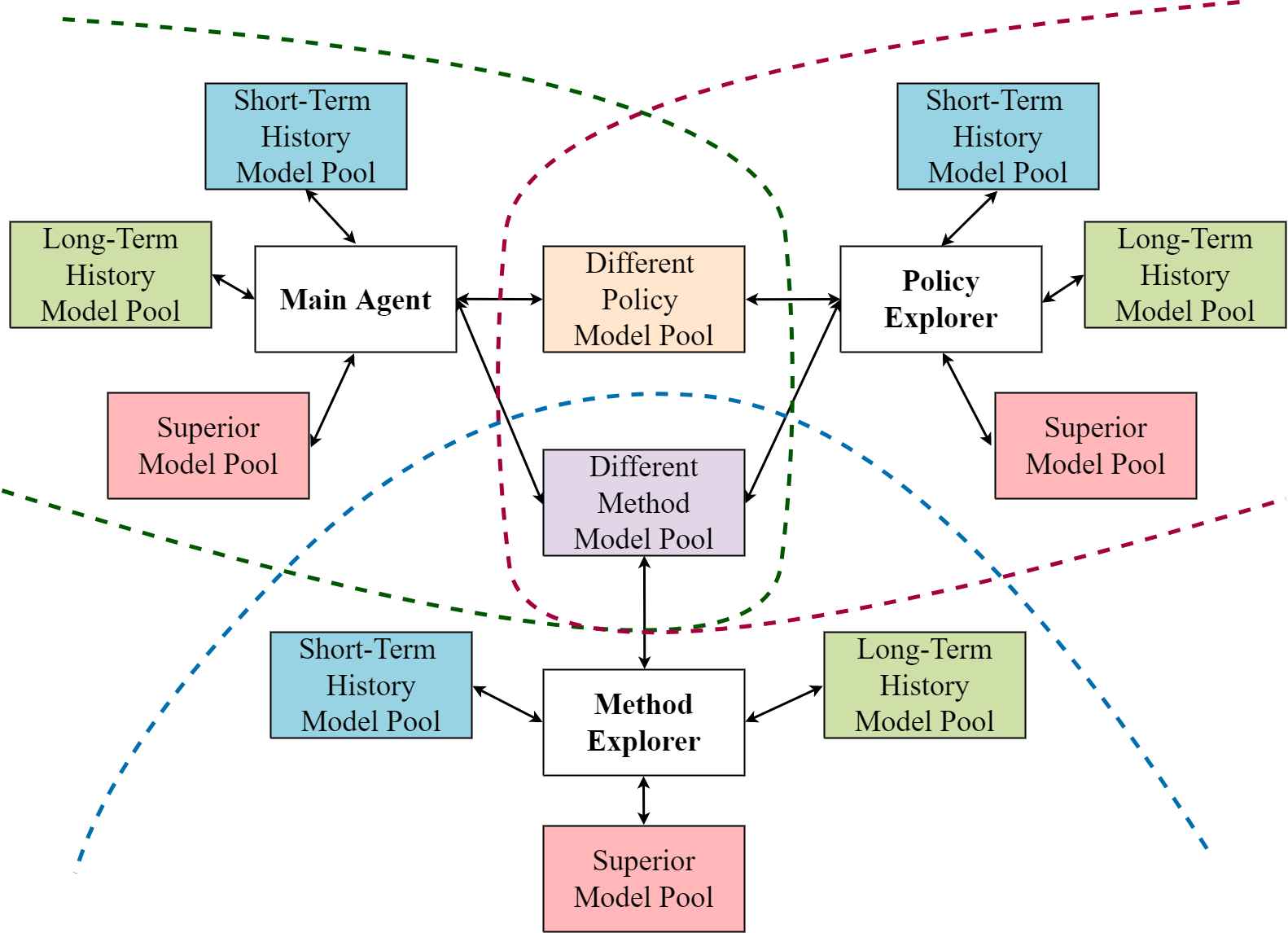}
	\caption{The diagram of the relation between the three kinds of AIs and their HMPs.}
	\label{fig:3}
\end{figure}

Increasing diversity among opponents and states can indirectly enhance the exploration of training model, thereby improving its generalization. To this end, five different history model pools (HMPs) were designed for the DIS, including the short-term history model pool (SHMP), long-term history model pool (LHMP), superior model pool (SMP), different policy model pool (DPMP), and different method model pool (DMMP). These HMPs form a complex structure that connects the three AIs, as illustrated in Figure \ref{fig:3}. Each of the AIs has its own SHMP, LHMP, and SMP. The SHMP stores short-term history models with a fixed capacity, replacing old models with new ones. The LHMP stores training models every 12 hours and retains all previously stored models. The SMP periodically stores superior models with good performance as determined by model evaluation processes, which is detailed in Section 3.3. The LHMP and SMP ensure that training AIs can defeat strategies seen in the past to prevent catastrophic forgetting \cite{ramasesh2022effect} or strategy collapse. The DPMP is shared between the main agent and policy explorer, while the DMMP is shared among all AIs. In addition to being stored in the SMP, the superior models of the policy explorer and main agent are also sent to the DPMP, and the superior models of all AIs are sent to the DMMP. These 5 HMPs are enough to store rich and excellent strategies. The samplers can select history models as opponents by sending model requests to the corresponding HMP, which then distributes models to the corresponding samplers of AIs.

\subsection{Training System}
\label{Section3.2}

\subsubsection{Training methods of Three AIs}
\label{Section3.2.1}
As previously described, it is anticipated that these three AIs will have distinct strategies; therefore, diverse models are employed to train their respective models. The main agent utilized the dual-clipping version \cite{ye2020mastering} of Proximal Policy Optimization (PPO) \cite{schulman2017proximal}, which is a popular actor-critic algorithm \cite{mnih2016asynchronous}. The dual-clipping can mitigate the training instability brought by the asynchronous update of the models on our distributed training framework. The importance ratio of policy in the PPO takes the form: $ r_{t} = \frac{\pi(a_t|s_t)}{\pi_{old}(a_t|s_t)} $, where $ s_t $ and $ a_t $ represent state and action at time step \textit{t}, respectively. $ \pi(a_t|s_t) $ and $ \pi_{old}(a_t|s_t) $ are the current and old policies, respectively. We denote $ r_t^c=clip(r_t,1-\varepsilon,1+\varepsilon) $, which is the clipped ratio in the PPO.

Hence, the policy objective of the agent is defined as: 
\begin{equation}
	\label{eq1}
	L_p = 
	\begin{cases}
		-\mathbb{\hat E}_t[max(min(r_{t} {\hat A}_t,r_t^c{\hat A}_t)), \eta{\hat A}_t] &  {\hat A}_t < 0 \\
		-\mathbb{\hat E}_t[min(r_{t} {\hat A}_t,r_t^c{\hat A}_t)] &  {\hat A}_t \geq 0 \\
	\end{cases}
\end{equation}
where $ -\mathbb{\hat E}[...] $ indicates the expectation over a finite batch of samples, $ A_t $ is the advantage of the agent at timestep \textit{t}, which is calculated via Generalized Advantage Estimation (GAE) \cite{schulman2015high}. $ \varepsilon $ and $ \eta $ are the original clip and the dual-clip PPO hyperparameters of agents, respectively.

The value objective of the agent in the PPO takes the form:
\begin{equation}
	\label{eq2}
	L_v = \mathbb{\hat E}_t[ ({ V}(s_t) -G_t )^{2}] 
\end{equation}
where $ G_t = V_{old}({s}_t) + {\hat A}_t $ is the returns of the agent, and $ V_{old}(s_t) $ is calculated by the old value function. 

The policy explorer, which is a variant of the main agent, has distinct changes to network input, reward settings, and training parameters. In this study, we primarily utilized three policy explorers: the first explorer augmented input with historical information; the second explorer employed goal clipping; the last explorer utilized ball possession reward. Incorporating historical information in the input can enable the model to remember previous information. Goal clipping was implemented to constrain the goal difference within the range of [-3, 3], thus preventing anomalous operations in situations where the AI falls behind by a significant score that rarely occurs during the training process. By constraining the goal difference to a certain range, the model can strive to win the game in most cases. Moreover, we employed the ball possession reward to encourage the model to retain the ball, which results in a more defensive strategy when leading the score.

To further enrich the strategies of the models in HMPs, the method explorer is trained by different DRL algorithms from the main agent. In this study, we employed Random Network Distillation (RND) \cite{burda2018exploration} and multi-agent PPO (MAPPO) \cite{guo2020joint} as the algorithms of method explorers. RND is an intrinsic reward-based method \cite{du2019liir,zheng2020can} that stimulates agent exploration by encouraging it to earn intrinsic rewards through searching unfamiliar states. We combined RND with Proximal Policy Optimization (PPO) to train a new AI with strategies distinct from those trained using classical PPO. The policy objective of RND-PPO is the same as that of PPO, except that the advantage of RND-PPO is the weighted sum of the original advantage and the intrinsic advantage calculated from the intrinsic rewards. The new advantage in the RND-PPO takes the form:
\begin{equation}
	\label{eq3}
	{\hat A}^{rnd}_t  = {\hat A}^e_t  + c^i {\hat A}^i_t 
\end{equation}
where $ {\hat A}^e_t $ and $ {\hat A}^i_t $ are the extrinsic (original) advantage and the intrinsic advantage, respectively. And $ c^i $ is the coefficient of intrinsic advantage. Hence, the policy objective of RND-PPO is denoted as $ L^{rnd}_p $ that is similar with the Eq.(\ref{eq1}), where the $ {\hat A}_t  $ is replaced by the $ {\hat A}^{rnd}_t  $.

The value objective of RND part is also the same as that of PPO. The only difference is that the returns of the agent are calculated by the intrinsic advantage, which is defined as:
\begin{equation}
	\label{eq4}
	L^{rnd}_v = \mathbb{\hat E}_t[ ({V^{rnd}}({s}_t) -G^{rnd}_t )^{2}] 
\end{equation}
where $ G^{rnd}_t = V^{rnd}_{old}({s}_t) + {\hat A}^i_t $ is the returns of the agent. 

Hence, the total loss of the RND-PPO can be represented as follows:
\begin{equation}
	\label{eq5}
	L^{rnd-ppo} =  L^{rnd}_p  + L_v  + L^{rnd}_v + L^{rnd}_{pre}
\end{equation}
where $ L^{rnd}_{pre} $ is the prediction loss for estimating the familiarity of the state, which is used to encourage the agent to explore the unfamiliar states. The weights in the RND loss part are omitted in this formula.

MAPPO is a multi-agent RL (MARL) \cite{zhang2021multi} algorithm that allows PPO to follow the Centralized Training and Decentralized Execution (CTDE) structure \cite{rashid2020monotonic}. Each policy of player in the MAPPO makes the decision based on their local information; while the value function will be trained by the global information. Hence, the policy objective of MAPPO can be denoted as $ L^{mappo}_{p} = \sum_1^{k=n}{L_p^k / n} $, where the $ L_p^k $ is the policy objective of kth player, \textit{n} is the number of players. The value objective of MAPPO is similar to that of PPO, while it uses the global information as the state.

In the preliminary experiments, we evaluated other algorithms that can be used for the method explorer, such as the Soft Actor-Critic (SAC) \cite{haarnoja2018soft}. Here, we compared the performance achieved by four established algorithms, PPO, RND-PPO, MAPPO, and SAC. Finally, we employed the RND-PPO and MAPPO as the algorithms of method explorer. The details of the comparison results for these four algorithms are provided in Appendix \ref{appendix:1}.

\subsubsection{Modeling and Reward Shaping}
\label{Section3.2.2}
The GRF environment offers two observation representations, namely the mini-map and vector information. The mini-map is a tensor with a shape of 4 × 72 × 96, while vector information is a one-dimensional tensor with a length of 115. Some previous explorations \cite{huang2021tikick} show that using the mini-map as the network input will severely reduce the sample efficiency and increase the number of model parameters. Therefore, we improved the original vector information to a new representation with a length of 214 and 370 as the network input for 5v5 training and 11v11 training, respectively. The enhanced vector information incorporates features such as controlling agent, ball, teammates, opponents, offside judgment, match status, yellow/red cards, etc. All methods in this study use the same network input, except for MAPPO, which breaks down the network input into local and global information for CTDE training. The default action set for each player, consisting of 19 actions, was utilized for the output. No action masks were used to expedite the training process, as hand-crafted action rules may restrict the range of model strategies. Appendix \ref{appendix:2} provides more information on the network input and output details.

The network structure has six dense layers for both the policy and value networks, and the policy and value networks share the first three layers. However, the network structures in those employed algorithms are slightly different, for example, the policy net and value net didn’t share layers for the MAPPO. The details of the network structure for all methods in this study are provided in Appendix \ref{appendix:3}. 

To solve the sparse reward issue \cite{hare2019dealing} in the GRF environment, our approach involves designing multiple rewards to guide the model's learning process, in addition to utilizing wins and goals as rewards. However, an overreliance on hand-crafted rewards during training may lead to reduced generalization and a limited set of strategies for the model. Therefore, the main agent only employed the most significant rewards, such as win, loss, goal, and ball possession change. Meanwhile, policy explorers are free to utilize various combinations of all designed rewards to capture models with distinct strategies, such as holding ball reward, secondary attack reward, etc. The details of the reward shaping for all AIs are provided in Appendix \ref{appendix:4}.

\subsubsection{Model Sampling Method}
\label{Section3.2.3}
\begin{figure}[h]
	\centering
	\includegraphics[scale=0.15]{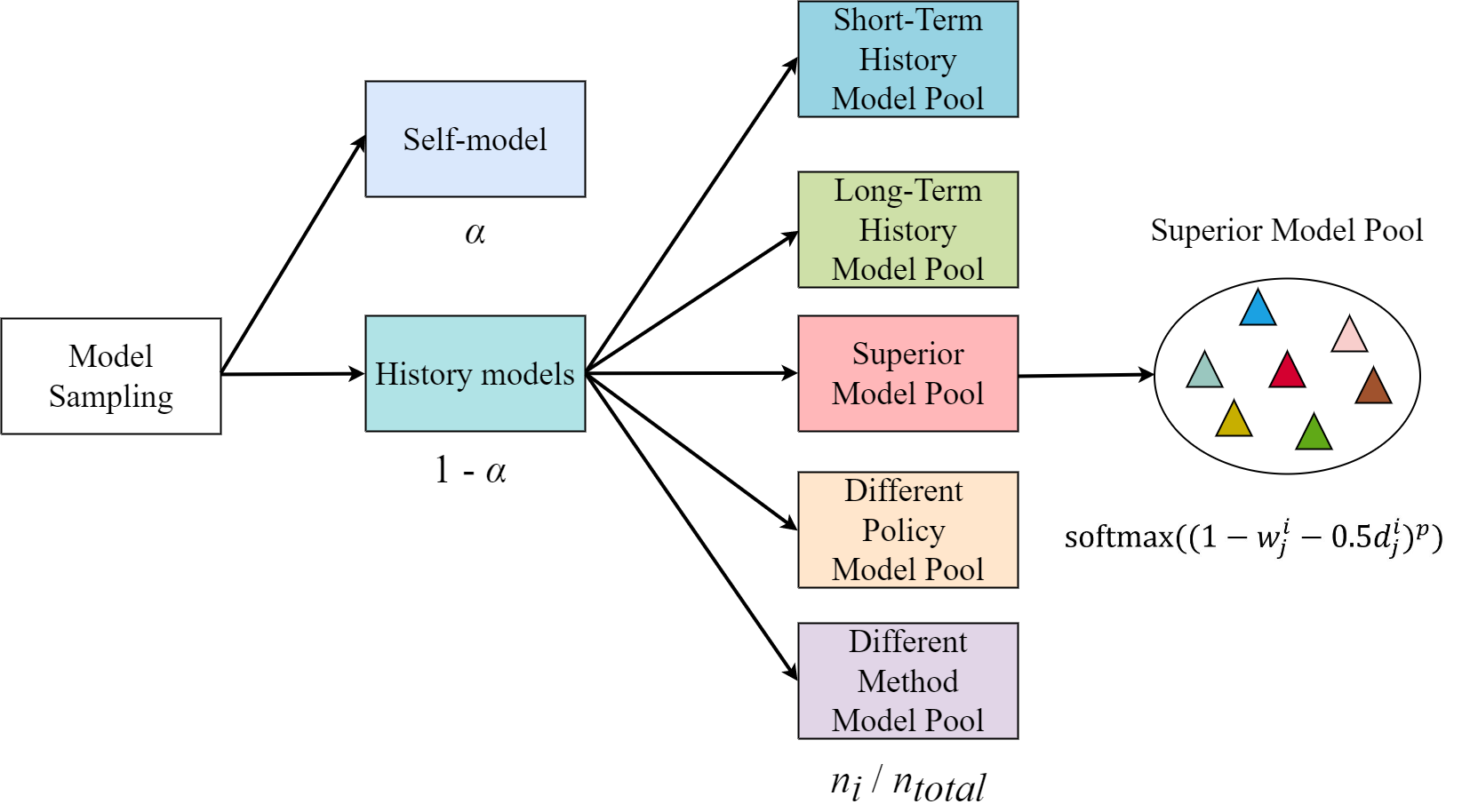}
	\caption{Example of model sampling process that the AI chooses a history model from the SMP, the sampling probability of each part is given below the corresponding diagram identifier.}
	\label{fig:4}
\end{figure}

During the training process, the samplers, which consist of three types of AI, select either the current self-model or historical models as their opponents. Due to limited training resources, playing against opponents with superior strategies can improve the training performance. Hence, we introduce a model sampling method (MSM) to improve the effectiveness of model sampling. We firstly assign the sampling probability of the self-model to $ \alpha $, so the probability of choosing the history models is $ (1-\alpha) $. As described in Section 3.1, each AI is equipped with five kinds of HMPs, with varying numbers of models in each pool. Therefore, we assign the model sampling probability for each HMP to be proportional to the number of models contained within it. So the sampling probability of the HMP \textit{i} is $ p_i=(1-\alpha)n_i⁄n_{total} $, where $ n_i $ is the number of history models in HMP \textit{i}, the $ n_{total} $ is the number of all history models that one AI can sample. For each kind of AIs, we use the following formula to determine the probability of sampling the history model \textit{j} in the HMP \textit{i}, which is noted by
\begin{equation}
	\label{eq6}
	P_j^{i}  = p_i\cdot softmax((1-w_j^i-0.5d_j^i)^p) 
\end{equation}
where $ w_j^i $ and $ d_j^i $ represent the win rate and draw rate of the current model against the history model $ j $, respectively. The parameter $ p $ controls the importance of the win rate and draw rate, which is set to 1 in this work. The softmax function is used to convert the importance values to probabilities, and a softmax temperature is employed to control the extent of this conversion. Figure \ref{fig:4} illustrates an example of the model sampling process, which shows how an AI selects a historical model from the SMP via the calculated sampling probability.

\subsubsection{Specific Scenario Training}
\label{Section3.2.4}
\begin{figure}[h]
	\centering
	\includegraphics[scale=0.18]{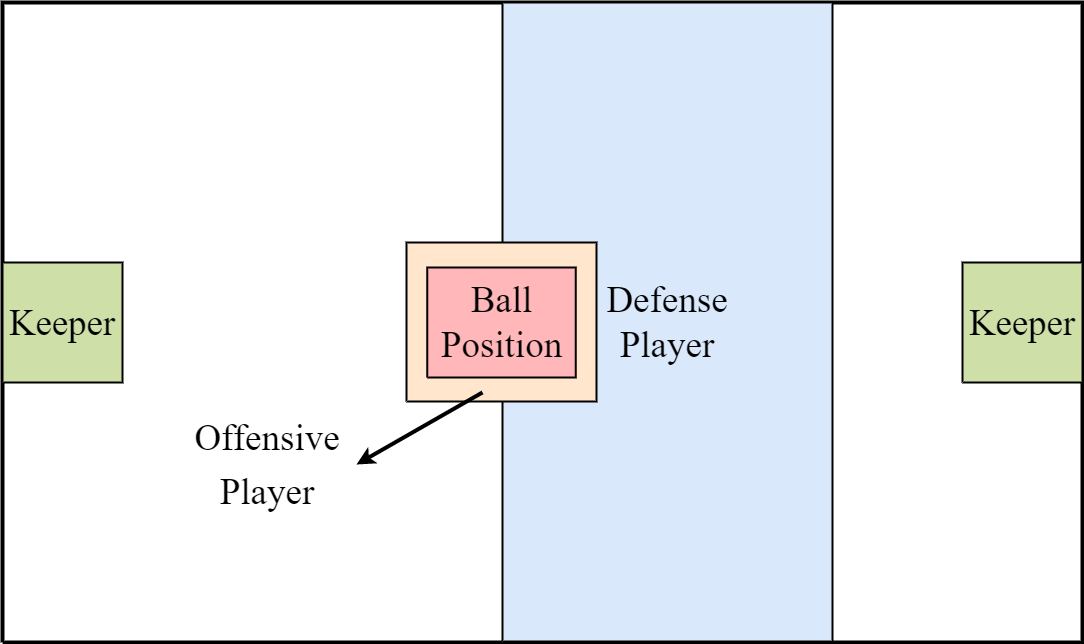}
	\caption{Diagram of the specific scenario of ``solo for 5v5 track''. In the scenario, one defense player will be randomly generated in the area of the blue box. The ball and the offense player will be randomly generated in the area of the red box and yellow box, respectively. Other players will be randomly generated throughout the field, and the goalkeepers will be presented in their fixed positions. }
	\label{fig:5}
\end{figure}

To further augment the diversity of states and increase the model's generalization, we devised specific scenarios to facilitate training. These scenarios incorporate frequent and crucial matchup situations, such as kick-off, corner kicks, free kicks, and attacks in the penalty area, among others. The specific scenario training (SST) enhances the model's strategy level in diverse states and improves its capacity to handle various special events. Each specific scenario lasts for a maximum of 512 steps. To ensure training variety, the ball and players' positions are randomly generated within a specific range in these scenarios. For instance, the ``solo for 5v5 track'' scenario is employed to train the agent's 1v1 situation handling abilities, as depicted in Figure \ref{fig:5}. During training, the models are randomly assigned to either the offensive or defensive side, allowing them to enhance their offensive capabilities in these scenarios while learning to defend themselves. A list of those specific scenarios we have designed is given in the Supplementary Files.

\subsection{Model Evaluation}
\label{Section3.3}
\begin{figure}[h]
	\centering
	\includegraphics[scale=0.14]{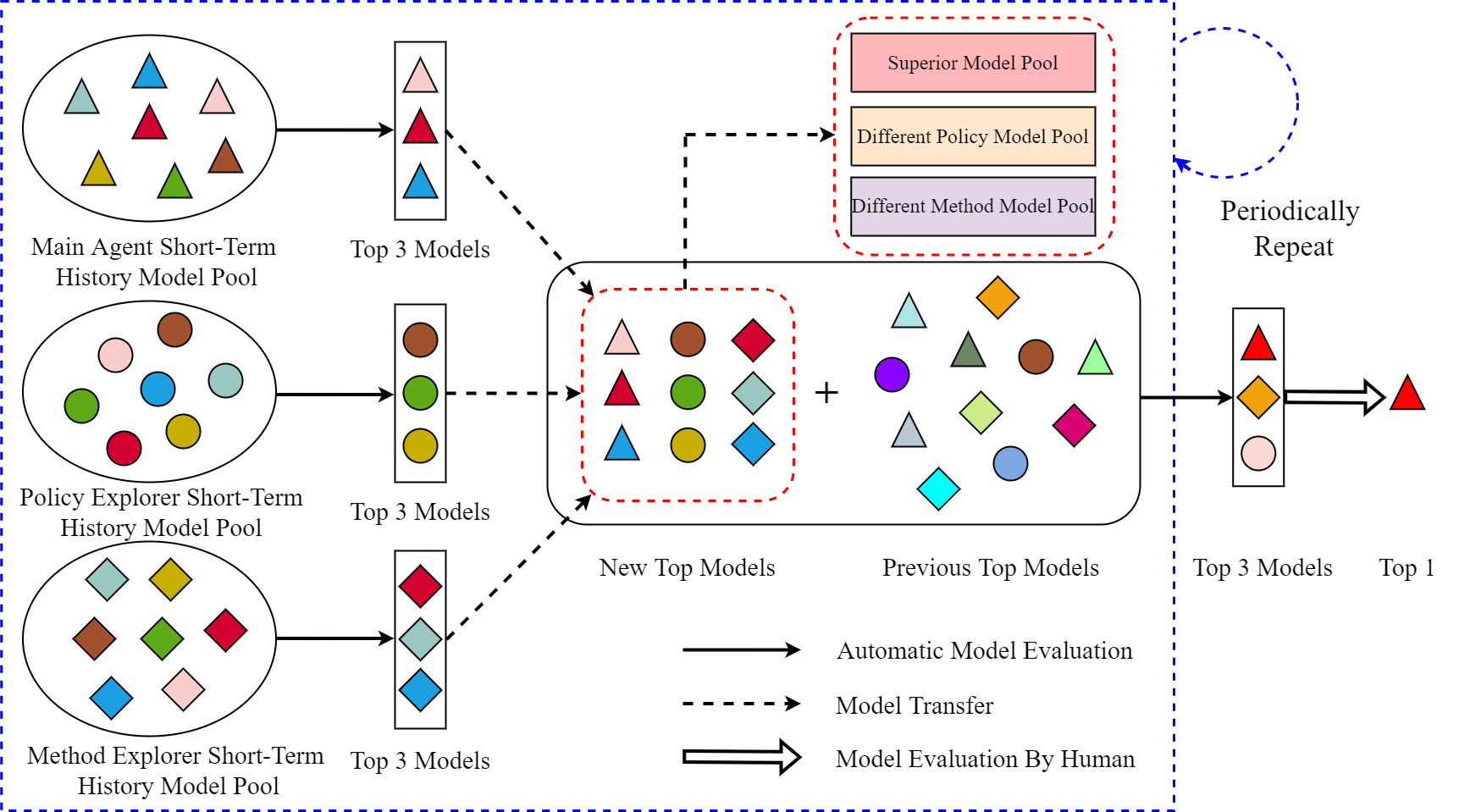}
	\caption{Illustration of the model evaluation process.}
	\label{fig:6}
\end{figure}

The quality of the model pool is an important factor for achieving a target AI. And the quality of model pool is directly determined by the models in it. Therefore, it is critical to select appropriate models through a scientific and reasonable model evaluation process. Our proposed model evaluation process is described in Figure \ref{fig:6}. During the training process, history models are periodically generated and stored in their corresponding SHMPs, as shown on the left side of the figure. The top three models, based on their Elo scores, are selected through a series of battles against the models from their own group. The new top models are then added to their corresponding SMPs, DPMPs, and DMMPs for further training. Additionally, the new top models are merged with the previous top model to obtain a new top model pool. We regularly evaluate the models in the top model pool by having them play against each other to assess their win rates and Elo scores. We rank the results and select the top three models from the top model pool, eliminating the lower-ranked models. This selection process is repeated during model training to expand the HMPs. Finally, we apply an additional human judgment process to the top three models of the top model pool to obtain the best model. The human judgment process considers the model's win rate and Elo score, as well as factors such as strategy style, average number of goals, and average ball possession.

\section{Experiments}
\label{Section:4}
\subsection{Training Process}
\label{Section:4.1}
During the training process, the main agent's training schemes, such as main hyperparameters and reward settings, remained unchanged. The policy explorer periodically replaced the training model with the current model of the main agent and then fine-tuned the model with different reward settings or hyperparameters. Two method explorers respectively used the RND-PPO and MAPPO algorithms to train the AIs. The details of the training hyperparameters are given in Appendix \ref{appendix:5}. Figure \ref{fig:7} gives the diagram of the performance change of the 5v5 model of main agent as the training time increases. It demonstrates that the performance of the model can be steadily improved with the DIS training iterations; and the improvement trends did not decrease significantly over time. The performance changes of the 11v11 model throughout the training process are presented in Appendix \ref{appendix:6}. Following training, the best-selected models achieved championship titles in all tracks of the IEEE CoG 2022 Football AI Competition. The training process cost 24 Nvidia 2080Ti GPUs and 9000 CPU cores, each type of AI in the DIS framework consumed one-third of the resources. The training process was implemented by Python 3.6 with Pytorch 1.6. Some game videos and goal highlights played by our proposed AI are available at: \url{https://sourl.cn/BCstgt}.
\begin{figure}[h]
	\centering
	\includegraphics[scale=0.054]{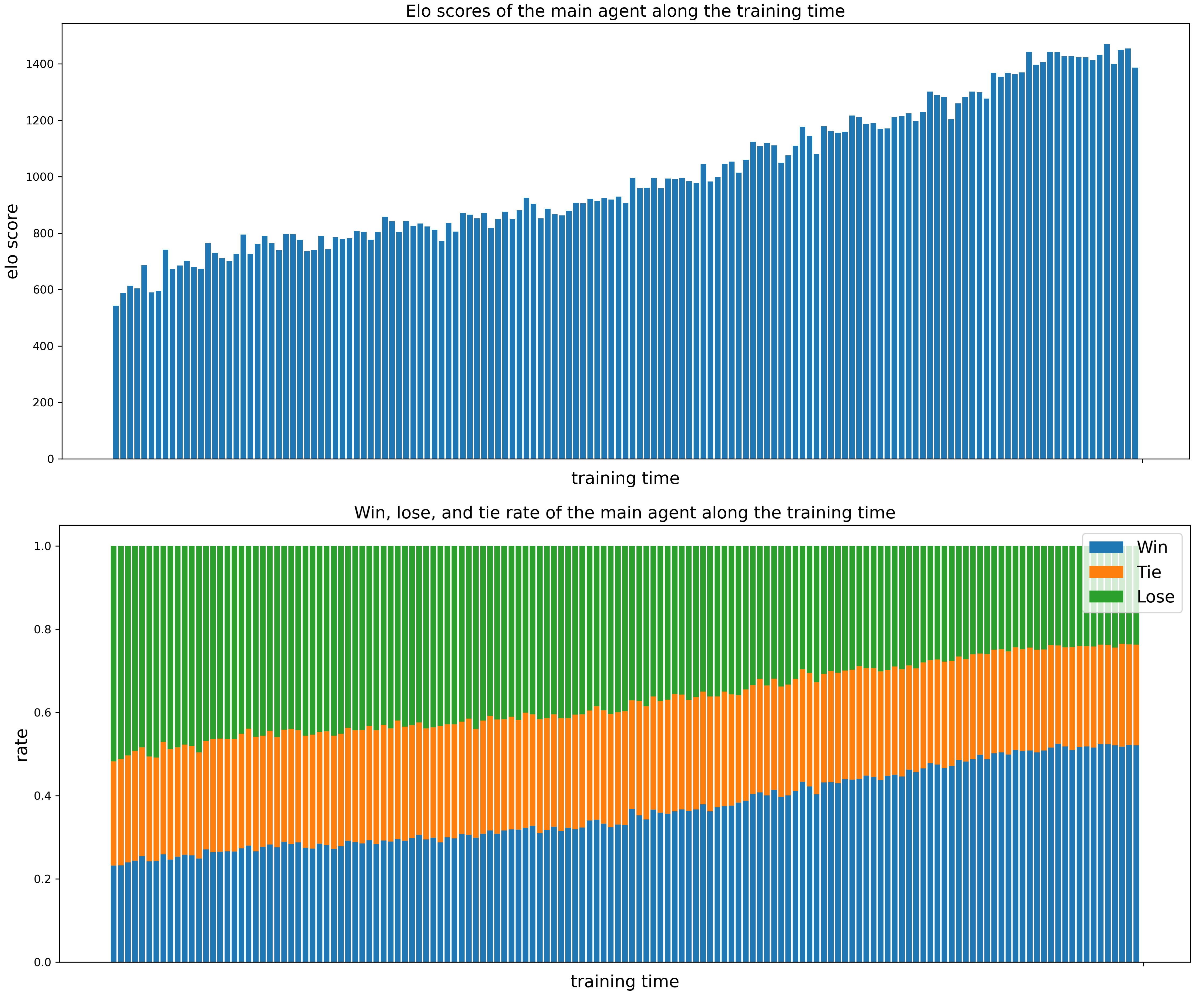}
	\caption{Diagram of the performance change of 5v5 model of main agent during training process, spanning 70 days from the start of training to the end of the competition. The results were obtained by fighting against all superior models in the training process. The first row depicts the steady improvement of the model's Elo values during the training process. The second row highlights the variation of the model's win rate, tie rate, and lose rate. It is clear that both the Elo value and win rate of the model have been increasing, and the growth trend has remained consistent throughout the training period.}
	\label{fig:7}
\end{figure}

\subsection{Behavior Analysis}
\label{Section:4.2}
\begin{figure}[h]
	\centering
	\includegraphics[scale=0.27]{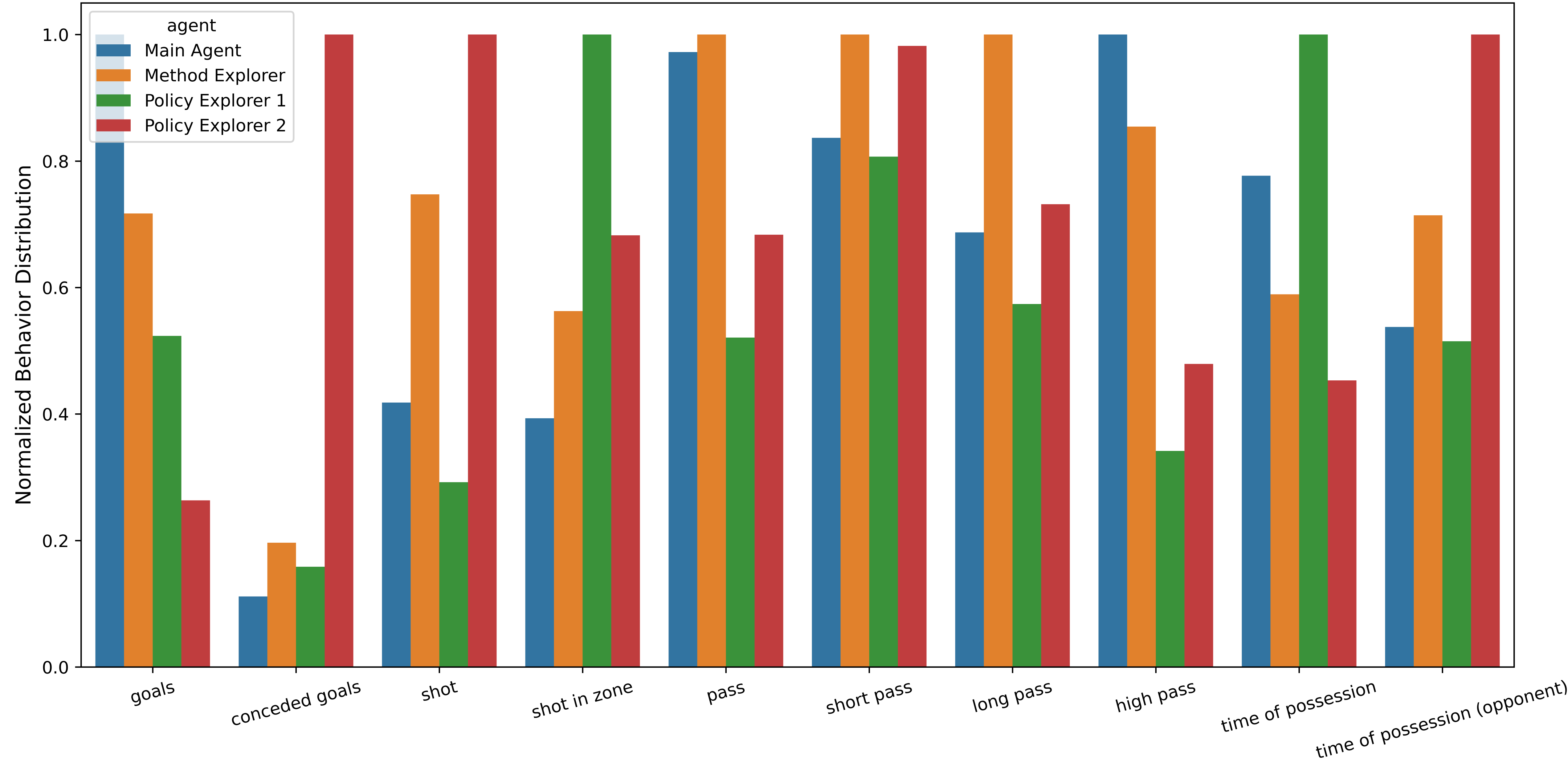}
	\caption{The normalized distribution of the ten behaviors with strategy-level significance for the main agent, policy explorer 1, policy explorer 2, and method explorer. It is noted that the main agent has the most goals, but its shot, pass, and possession are not the highest, which suggests the possibility of a more evenly distributed behavior and diverse strategies. Policy explorer 1 exhibits the highest possession time, the lowest willingness to pass, and the lowest possession time of the opponent. These results indicate that the inclusion of ball possession reward proves to be effective, making it an AI that prefers to carry the ball and has a higher defensive strength. Policy explorer 2 manifests a high willingness to shot and short pass, but its behaviors such as shot in zone, long pass, and high pass are less, leading to a lower score. Method explorer displays more of each behavior, but its score falls short of the main agent's score. This observation reflects a greater accentuation on exploration, aligning with the objective of choosing the RND-PPO as its core algorithm.}
	\label{fig:8}
\end{figure}

In addition to having high strength, our expectation is for the DIS-trained AIs to exhibit different strategies. Thus, we designed a test to evaluate the behaviors of the AIs as a means of expressing their strategies. The test involved a main agent, two policy explorers, and a method explorer. The policy explorers consisted of two models trained with ball possession reward and historical information incorporated into the network input, respectively. The method explorer was optimized with the PPO-RND algorithm. Each of these four AIs played against an identical model for one hundred rounds, and ten behaviors with strategy-level significance were counted. The normalized distributions of the ten behaviors for the four different AIs are presented in Figure \ref{fig:8}. The behavior distributions of these AIs exhibit significant differences, indicating diverse strategies. Playing against these AIs can improve the generalization of the main agent, enabling it to handle more states and enrich the diversity of AI strategies.

\begin{figure}[h]
	\centering
	\includegraphics[scale=0.08]{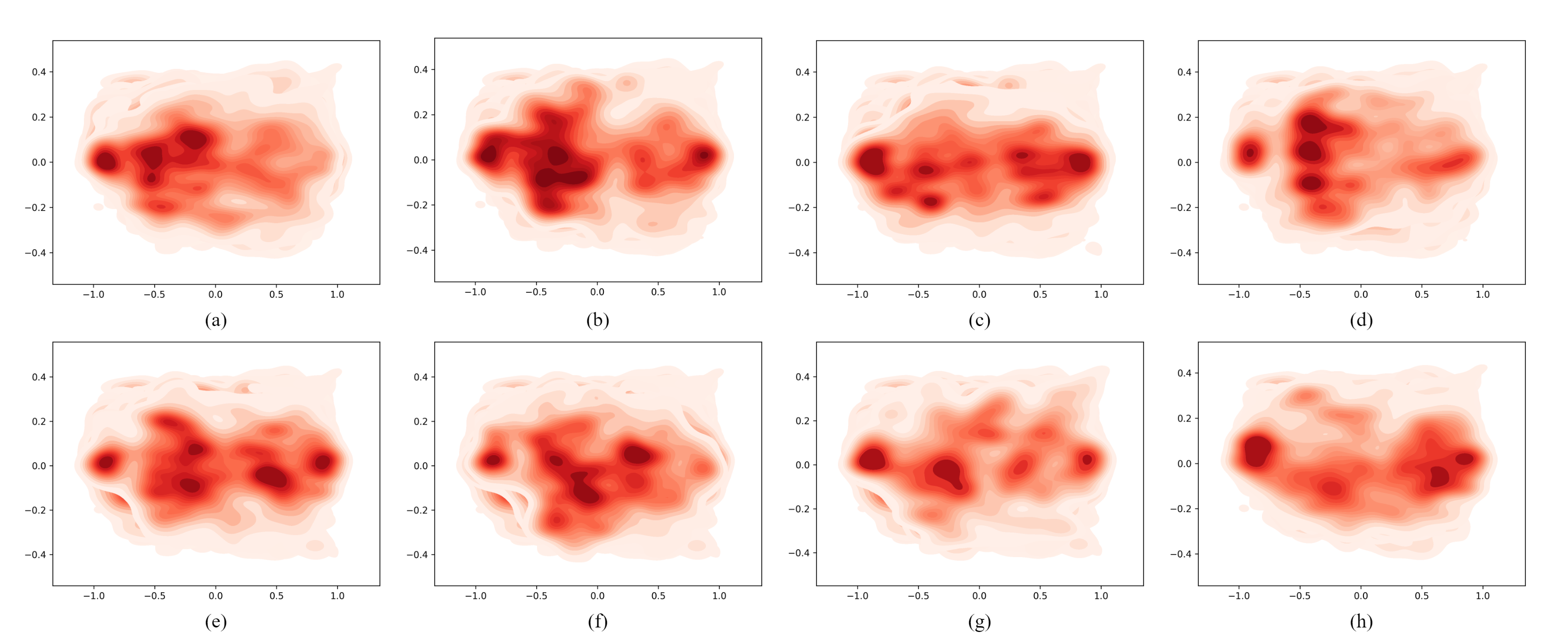}
	\caption{Heatmaps of player positions for final 5v5 model against eight history models from different training periods of different AIs. When confronting models (a) and (d), the players of the final model are primarily concentrated in the backcourt, indicating a defensive inclination. When playing against models (b), (c), and (h), the players are concentrated in both the frontcourt and backcourt, pointing to the final model's ability to balance both defense and offense. Furthermore, when facing models (e), (f), and (g), the players of the final model tend to move towards the midfield, which can emphasize midfield control. }
	\label{fig:9}
\end{figure}

Next, we conducted an analysis of player position distribution in the final 5v5 model when pitted against eight different history models. These models were selected from various training periods of diverse AIs, each with distinct strategies. The final 5v5 model played ten games against each history model, and the corresponding heatmaps of player positions are depicted in Figure \ref{fig:9}. The heatmaps offer a clear visualization of the concentration of player positions on the field during the game. The results indicate that the final AI exhibited a noticeable difference in position distribution when matched against different opponents, indicating that our proposed model utilizes varied strategies against different opponents. The behavior analysis results demonstrate the rich diversity of strategies and the superior generalization ability of the trained AI.

\subsection{Ablations}
\label{Section:4.3}
\begin{figure}[h]
	\centering
	\includegraphics[scale=0.44]{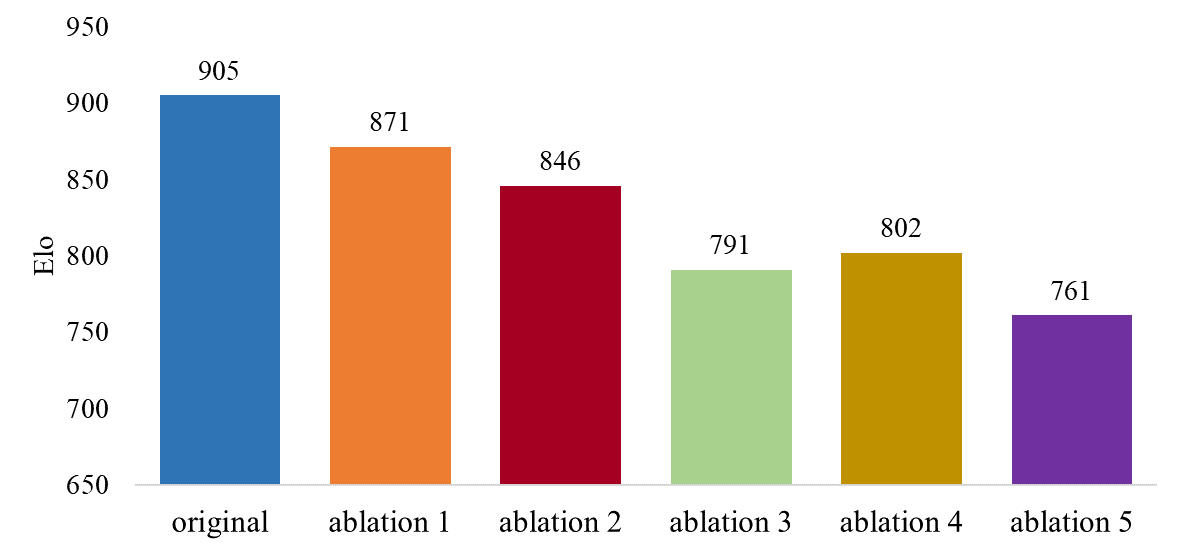}
	\caption{The Elo values of the models with different key modules of the DIS.}
	\label{fig:10}
\end{figure}

Two ablation tests were conducted to investigate the impact of key modules in the proposed method. As detailed in Section 3.1, the SHMP, LHMP, SMP, DPMP, DMMP are the pivotal modules in the DIS. Therefore, the first set of tests trained an original model and five ablation models to validate the effectiveness of these modules. The composition of these models is illustrated in Table \ref{table:1}. All test models were trained from a starting model that had been trained with the proposed method for one month. The training mechanism of the modules in these models followed that of the proposed method. The evaluation criterion for these models was the Elo values after 48 hours of training, as shown in Figure \ref{fig:10}. The results indicate that the use of SHMP, SMP, DPMP, and DMMP enhances the performance of the AI. However, the model in ablation 4, which removed the LHMP, outperformed the model in ablation 3, which included both SHMP and LHMP. This result may be attributed to the fact that the LHMP contains some early training models, which are weak but consume training resources, thereby reducing performance. Nevertheless, we retain the LHMP to improve the AI's generalization ability against weaker models. For instance, we observed that AI that only advances with the ball sometimes has an advantage over models with higher Elo scores. Additionally, the comparison between ablation 4 and ablation 5 demonstrates that the addition of the SHMP module is superior to utilizing only the self-play training, which further confirms the usefulness of the HMP in self-play training.

\begin{table}[h]
	\centering
	\begin{tabular}{ll}
		\toprule
		Models  &  Modules                              \\
		\midrule
		Original       &    SHMP + LHMP + SMP + DPMP + DMMP       \\
		Ablation 1     &    SHMP + LHMP + SMP + DPMP      \\
		Ablation 2     &    SHMP + LHMP + SMP     \\
		Ablation 3     &    SHMP + LHMP     \\
		Ablation 4     &    SHMP     \\
		Ablation 5     &    Only self-play training     \\
		\bottomrule
	\end{tabular}
	\caption{The components of trained models in the first ablation test.}
	\label{table:1}
\end{table}

In the second group, we conducted experiments to evaluate the impact of four factors on the performance of our method, including MSM, SST, built-in AI as opponents, and action masks. Hence, four models were trained to assess the effectiveness of these factors, using the same starting model as in the first group. The first ablation model employed random sampling instead of MSM for the history models. The second model removed the SST module from our proposed method. The third model incorporated a 10\% probability of playing against built-in AI during training. The last model utilized a designed action mask to prevent meaningless actions, such as a shot at the backfield. The details of the designed action mask can be found in Appendix \ref{appendix:7}. Figure \ref{fig:11} depicts the Elo values of these comparison models trained for 48 hours. The results demonstrate that the proposed MSM enhances training performance. The SST effectively improves training by reinforcing the tactics of crucial scenarios. Notably, incorporating built-in AI as opponents during training significantly reduced model performance. Upon analyzing the corresponding game replays, we observed that playing against built-in AI caused the model to overfit simple strategies, such as scoring from the side pass, which did not provide an advantage against other models. Another factor worth examining is whether adding an action mask improves AI performance in the GRF environment. The original model was trained without an action mask, and in the ablation test, we fine-tuned the starting model by adding an action mask. However, the results showed a slight reduction in the Elo value after adding the action mask due to the learned strategies being limited by the action mask. For instance, shooting from the backfield has the effect of clearance. Additionally, the starting model was trained without an action mask, which may make it unsuitable for training with an action mask. Each test used 1000 samplers during the training process.

\begin{figure}[h]
	\centering
	\includegraphics[scale=0.36]{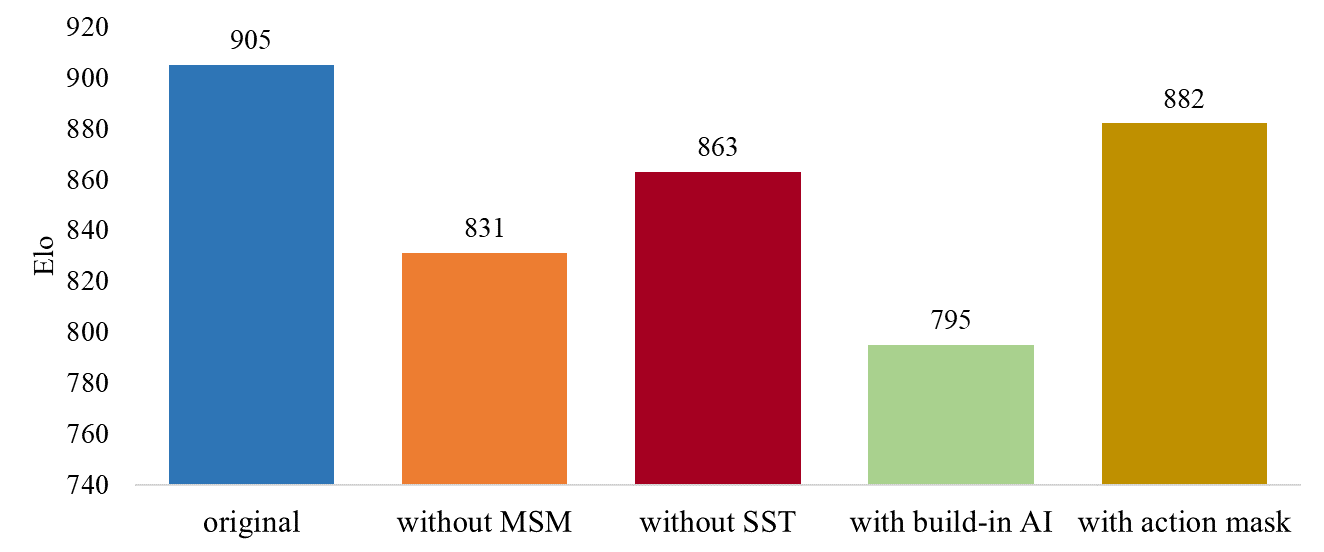}
	\caption{The Elo values of the models with different factors.}
	\label{fig:11}
\end{figure}

\section{Conclusion}
\label{Section:5}
This paper proposes a novel DRL training framework that simultaneously trains three distinct AIs. These AIs share their respective historical model pools, creating a complex interconnected model pool structure. This framework enables the AIs to combat various opponents, thereby enhancing their strength and strategy richness. The model evaluation and screening scheme proposed in this paper ensure that the models with satisfactory performance are incorporated into the model pool or serve as the target model. The proposed framework provides trained AIs with diverse strategies, better generalization, and a high level of performance. In a GRF competition, the trained AIs won the 5v5 and 11v11 tracks, marking the first time an AI achieved excellent performance in a full football game without using human data. Behavior analysis and ablation experiments confirmed that the trained AIs had rich strategies, and the framework's designed modules benefited the training process. Given that most of the modules in the framework are decoupled from the specific environment, such as the HMP structure and the algorithms employed by different agents, this framework can be applied to training RL-based AIs in most adversarial environments.

\bibliographystyle{named}
\bibliography{ijcai22}

\begin{thebibliography}{}

\bibitem[\protect\citeauthoryear{Balduzzi \bgroup \em et al.\egroup
  }{2019}]{balduzzi2019open}
David Balduzzi, Marta Garnelo, Yoram Bachrach, Wojciech Czarnecki, Julien
  Perolat, Max Jaderberg, and Thore Graepel.
\newblock Open-ended learning in symmetric zero-sum games.
\newblock In {\em International Conference on Machine Learning}, pages
  434--443. PMLR, 2019.

\bibitem[\protect\citeauthoryear{Berner \bgroup \em et al.\egroup
  }{2019}]{berner2019dota}
Christopher Berner, Greg Brockman, Brooke Chan, Vicki Cheung, Przemys{\l}aw
  Dkebiak, Christy Dennison, David Farhi, Quirin Fischer, Shariq Hashme, Chris
  Hesse, et~al.
\newblock Dota 2 with large scale deep reinforcement learning.
\newblock {\em arXiv preprint arXiv:1912.06680}, 2019.

\bibitem[\protect\citeauthoryear{Burda \bgroup \em et al.\egroup
  }{2018}]{burda2018exploration}
Yuri Burda, Harrison Edwards, Amos Storkey, and Oleg Klimov.
\newblock Exploration by random network distillation.
\newblock {\em arXiv preprint arXiv:1810.12894}, 2018.

\bibitem[\protect\citeauthoryear{de Woillemont \bgroup \em et al.\egroup
  }{2021}]{de2021configurable}
Pierre Le~Pelletier de~Woillemont, Remi Labory, and Vincent Corruble.
\newblock Configurable agent with reward as input: a play-style continuum
  generation.
\newblock In {\em 2021 IEEE Conference on Games (CoG)}, pages 1--8. IEEE, 2021.

\bibitem[\protect\citeauthoryear{Du \bgroup \em et al.\egroup
  }{2019}]{du2019liir}
Yali Du, Lei Han, Meng Fang, Ji~Liu, Tianhong Dai, and Dacheng Tao.
\newblock Liir: Learning individual intrinsic reward in multi-agent
  reinforcement learning.
\newblock {\em Advances in Neural Information Processing Systems}, 32, 2019.

\bibitem[\protect\citeauthoryear{Gleave \bgroup \em et al.\egroup
  }{2019}]{gleave2019adversarial}
Adam Gleave, Michael Dennis, Cody Wild, Neel Kant, Sergey Levine, and Stuart
  Russell.
\newblock Adversarial policies: Attacking deep reinforcement learning.
\newblock In {\em International Conference on Learning Representations}, 2019.

\bibitem[\protect\citeauthoryear{Guo \bgroup \em et al.\egroup
  }{2020}]{guo2020joint}
Delin Guo, Lan Tang, Xinggan Zhang, and Ying-Chang Liang.
\newblock Joint optimization of handover control and power allocation based on
  multi-agent deep reinforcement learning.
\newblock {\em IEEE Transactions on Vehicular Technology}, 69(11):13124--13138,
  2020.

\bibitem[\protect\citeauthoryear{Guss \bgroup \em et al.\egroup
  }{2021}]{guss2021minerl}
William~H Guss, Mario~Ynocente Castro, Sam Devlin, Brandon Houghton,
  Noboru~Sean Kuno, Crissman Loomis, Stephanie Milani, Sharada Mohanty, Keisuke
  Nakata, Ruslan Salakhutdinov, et~al.
\newblock The minerl 2020 competition on sample efficient reinforcement
  learning using human priors.
\newblock {\em arXiv preprint arXiv:2101.11071}, 2021.

\bibitem[\protect\citeauthoryear{Haarnoja \bgroup \em et al.\egroup
  }{2018}]{haarnoja2018soft}
Tuomas Haarnoja, Aurick Zhou, Pieter Abbeel, and Sergey Levine.
\newblock Soft actor-critic: Off-policy maximum entropy deep reinforcement
  learning with a stochastic actor.
\newblock In {\em International conference on machine learning}, pages
  1861--1870. PMLR, 2018.

\bibitem[\protect\citeauthoryear{Hare}{2019}]{hare2019dealing}
Joshua Hare.
\newblock Dealing with sparse rewards in reinforcement learning.
\newblock {\em arXiv preprint arXiv:1910.09281}, 2019.

\bibitem[\protect\citeauthoryear{Huang \bgroup \em et al.\egroup
  }{2021}]{huang2021tikick}
Shiyu Huang, Wenze Chen, Longfei Zhang, Shizhen Xu, Ziyang Li, Fengming Zhu,
  Deheng Ye, Ting Chen, and Jun Zhu.
\newblock Tikick: towards playing multi-agent football full games from
  single-agent demonstrations.
\newblock {\em arXiv preprint arXiv:2110.04507}, 2021.

\bibitem[\protect\citeauthoryear{Joyce}{2011}]{joyce2011kullback}
James~M Joyce.
\newblock Kullback-leibler divergence.
\newblock In {\em International encyclopedia of statistical science}, pages
  720--722. Springer, 2011.

\bibitem[\protect\citeauthoryear{Kurach \bgroup \em et al.\egroup
  }{2020}]{kurach2020google}
Karol Kurach, Anton Raichuk, Piotr Stanczyk, Michal Zajkac, Olivier Bachem,
  Lasse Espeholt, Carlos Riquelme, Damien Vincent, Marcin Michalski, Olivier
  Bousquet, et~al.
\newblock Google research football: A novel reinforcement learning environment.
\newblock In {\em Proceedings of the AAAI Conference on Artificial
  Intelligence}, volume~34, pages 4501--4510, 2020.

\bibitem[\protect\citeauthoryear{Li \bgroup \em et al.\egroup
  }{2020}]{li2020suphx}
Junjie Li, Sotetsu Koyamada, Qiwei Ye, Guoqing Liu, Chao Wang, Ruihan Yang,
  Li~Zhao, Tao Qin, Tie-Yan Liu, and Hsiao-Wuen Hon.
\newblock Suphx: Mastering mahjong with deep reinforcement learning.
\newblock {\em arXiv preprint arXiv:2003.13590}, 2020.

\bibitem[\protect\citeauthoryear{Mnih \bgroup \em et al.\egroup
  }{2016}]{mnih2016asynchronous}
Volodymyr Mnih, Adria~Puigdomenech Badia, Mehdi Mirza, Alex Graves, Timothy
  Lillicrap, Tim Harley, David Silver, and Koray Kavukcuoglu.
\newblock Asynchronous methods for deep reinforcement learning.
\newblock In {\em International conference on machine learning}, pages
  1928--1937. PMLR, 2016.

\bibitem[\protect\citeauthoryear{Mysore \bgroup \em et al.\egroup
  }{2022}]{mysore2023multi}
Siddharth Mysore, George Cheng, Yunqi Zhao, Kate Saenko, and Meng Wu.
\newblock Multi-critic actor learning: Teaching rl policies to act with style.
\newblock In {\em International Conference on Learning Representations}, 2022.

\bibitem[\protect\citeauthoryear{Ramasesh \bgroup \em et al.\egroup
  }{2022}]{ramasesh2022effect}
Vinay~Venkatesh Ramasesh, Aitor Lewkowycz, and Ethan Dyer.
\newblock Effect of scale on catastrophic forgetting in neural networks.
\newblock In {\em International Conference on Learning Representations}, 2022.

\bibitem[\protect\citeauthoryear{Rashid \bgroup \em et al.\egroup
  }{2020}]{rashid2020monotonic}
Tabish Rashid, Mikayel Samvelyan, Christian~Schroeder De~Witt, Gregory
  Farquhar, Jakob Foerster, and Shimon Whiteson.
\newblock Monotonic value function factorisation for deep multi-agent
  reinforcement learning.
\newblock {\em The Journal of Machine Learning Research}, 21(1):7234--7284,
  2020.

\bibitem[\protect\citeauthoryear{Risi and Preuss}{2020}]{risi2020chess}
Sebastian Risi and Mike Preuss.
\newblock From chess and atari to starcraft and beyond: How game ai is driving
  the world of ai.
\newblock {\em KI-Kunstliche Intelligenz}, 34:7--17, 2020.

\bibitem[\protect\citeauthoryear{Schulman \bgroup \em et al.\egroup
  }{2015}]{schulman2015high}
John Schulman, Philipp Moritz, Sergey Levine, Michael Jordan, and Pieter
  Abbeel.
\newblock High-dimensional continuous control using generalized advantage
  estimation.
\newblock {\em arXiv preprint arXiv:1506.02438}, 2015.

\bibitem[\protect\citeauthoryear{Schulman \bgroup \em et al.\egroup
  }{2017}]{schulman2017proximal}
John Schulman, Filip Wolski, Prafulla Dhariwal, Alec Radford, and Oleg Klimov.
\newblock Proximal policy optimization algorithms.
\newblock {\em arXiv preprint arXiv:1707.06347}, 2017.

\bibitem[\protect\citeauthoryear{Silver \bgroup \em et al.\egroup
  }{2016}]{silver2016mastering}
David Silver, Aja Huang, Chris~J Maddison, Arthur Guez, Laurent Sifre, George
  Van Den~Driessche, Julian Schrittwieser, Ioannis Antonoglou, Veda
  Panneershelvam, Marc Lanctot, et~al.
\newblock Mastering the game of go with deep neural networks and tree search.
\newblock {\em nature}, 529(7587):484--489, 2016.

\bibitem[\protect\citeauthoryear{Silver \bgroup \em et al.\egroup
  }{2018}]{silver2018general}
David Silver, Thomas Hubert, Julian Schrittwieser, Ioannis Antonoglou, Matthew
  Lai, Arthur Guez, Marc Lanctot, Laurent Sifre, Dharshan Kumaran, Thore
  Graepel, et~al.
\newblock A general reinforcement learning algorithm that masters chess, shogi,
  and go through self-play.
\newblock {\em Science}, 362(6419):1140--1144, 2018.

\bibitem[\protect\citeauthoryear{Tian \bgroup \em et al.\egroup
  }{2019}]{tian2019elf}
Yuandong Tian, Jerry Ma, Qucheng Gong, Shubho Sengupta, Zhuoyuan Chen, James
  Pinkerton, and Larry Zitnick.
\newblock Elf opengo: An analysis and open reimplementation of alphazero.
\newblock In {\em International conference on machine learning}, pages
  6244--6253. PMLR, 2019.

\bibitem[\protect\citeauthoryear{Vinyals \bgroup \em et al.\egroup
  }{2019}]{vinyals2019grandmaster}
Oriol Vinyals, Igor Babuschkin, Wojciech~M Czarnecki, Micha{\"e}l Mathieu,
  Andrew Dudzik, Junyoung Chung, David~H Choi, Richard Powell, Timo Ewalds,
  Petko Georgiev, et~al.
\newblock Grandmaster level in starcraft ii using multi-agent reinforcement
  learning.
\newblock {\em Nature}, 575(7782):350--354, 2019.

\bibitem[\protect\citeauthoryear{Vries}{2001}]{albers2001elo}
Han~de Vries.
\newblock Elo-rating as a tool in the sequential estimation of dominance
  strengths.
\newblock {\em Animal Behaviour}, pages 489--495, 2001.

\bibitem[\protect\citeauthoryear{Yannakakis and
  Togelius}{2018}]{yannakakis2018artificial}
Georgios~N Yannakakis and Julian Togelius.
\newblock {\em Artificial intelligence and games}, volume~2.
\newblock Springer, 2018.

\bibitem[\protect\citeauthoryear{Ye \bgroup \em et al.\egroup
  }{2020}]{ye2020mastering}
Deheng Ye, Zhao Liu, Mingfei Sun, Bei Shi, Peilin Zhao, Hao Wu, Hongsheng Yu,
  Shaojie Yang, Xipeng Wu, Qingwei Guo, et~al.
\newblock Mastering complex control in moba games with deep reinforcement
  learning.
\newblock In {\em Proceedings of the AAAI Conference on Artificial
  Intelligence}, volume~34, pages 6672--6679, 2020.

\bibitem[\protect\citeauthoryear{Zhang \bgroup \em et al.\egroup
  }{2021}]{zhang2021multi}
Kaiqing Zhang, Zhuoran Yang, and Tamer Ba{\c{s}}ar.
\newblock Multi-agent reinforcement learning: A selective overview of theories
  and algorithms.
\newblock {\em Handbook of reinforcement learning and control}, pages 321--384,
  2021.

\bibitem[\protect\citeauthoryear{Zheng \bgroup \em et al.\egroup
  }{2020}]{zheng2020can}
Zeyu Zheng, Junhyuk Oh, Matteo Hessel, Zhongwen Xu, Manuel Kroiss, Hado
  Van~Hasselt, David Silver, and Satinder Singh.
\newblock What can learned intrinsic rewards capture?
\newblock In {\em International Conference on Machine Learning}, pages
  11436--11446. PMLR, 2020.

\end{thebibliography}

\clearpage
\appendix

\section{Performance of Four Algorithms}
\label{appendix:1}

\begin{figure}[h]
	\centering
	\includegraphics[scale=0.34]{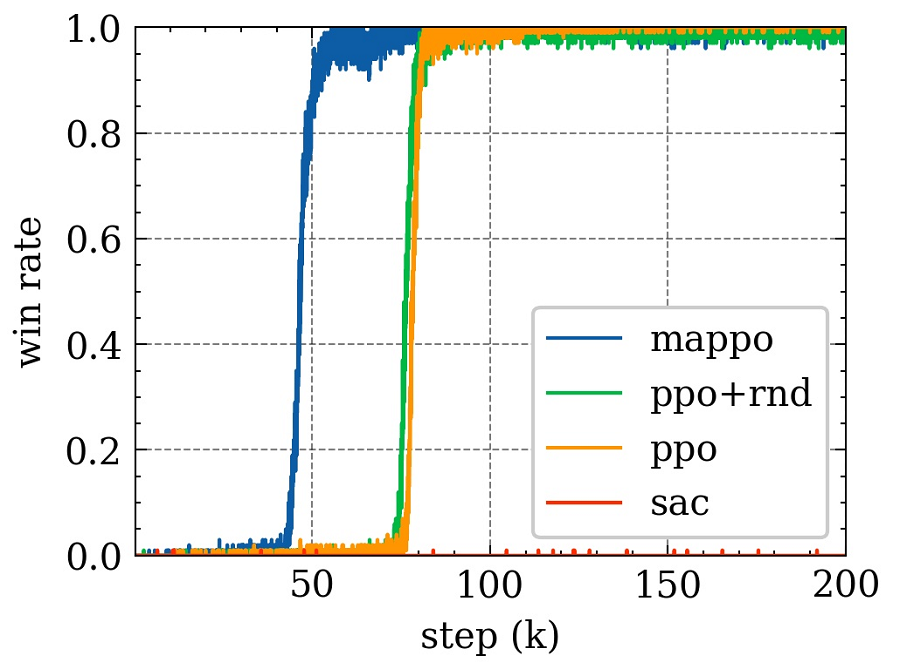}
	\caption{The win rate against the build-in AI for four different algorithms during the training process.}
	\label{fig:12}
\end{figure}

To compare the efficacy of different algorithms in the GRF environment and select the most effective ones for method explorers, we implemented four currently popular algorithms, namely, PPO, RND-PPO, MAPPO, and SAC. Using the 5v5 scenario as the test platform, we trained the model from scratch by competing against the built-in AI. The results, presented in Figure \ref{fig:12}, indicate that the MAPPO model exhibits the fastest convergence, while the RND-PPO exhibits convergence speeds similar to the PPO. However, SAC was unable to learn an effective strategy to defeat the built-in AI within the limited steps. As a method with CTDE architecture, MAPPO attempts to address the non-stable issue in multi-agent training. However, the centralized value function may cause problems of redundant information and dimensional explosion. Furthermore, the training requirements of MAPPO and RND-PPO algorithms are much greater than that of PPO. Therefore, we have selected PPO as the core algorithm of the main agent, while MAPPO and RND-PPO have been used for method explorers. SAC are not employed in this study.

\section{Details of Input Features and Output Actions}
\label{appendix:2}

Table \ref{table:2} presents the vector feature content utilized for 5v5 and 11v11 training. The feature content of the controlling player includes its information of position, direction, speed, role, tired factor, etc. The ball state includes the ball’s position, direction, speed, distance to the controlling player, owner, etc. The feature content for teammates and opponents is similar to that of the controlling player, but lacks a one-hot player role indicator. The closest teammate and opponent refer to the information of teammate and opponent closest to the controlling player, respectively. The available actions feature content is a multi-hot indicator, wherein 1/0 denotes if the corresponding action is available/unavailable for the controlling player in the current state. The match state records the time left, scores, game mode, etc. The offside judgement is used to indicate whether the players are in offside position. The feature content of yellow/red cards indicate whether the players have a yellow or red card. The feature content of sticky action is the indicator whether the players are in the particular state, such as the dribbling and sprinting. The same features in 5v5 and 11v11 training have different lengths due to the varying number of players. The default action set of the GRF environment, which consists of 19 actions (idle, left, top left, top, top right, right, bottom right, bottom, bottom left, long pass, high pass, short pass, shot, sprint, release direction, release sprint, sliding, dribble, and release dribble), is employed for the output form.

\begin{table}[h]
	\centering
	\begin{tabular}{lcc}
		\toprule
		Feature content  &  Length in 5v5   &  Length in 11v11                         \\
		\midrule
		Controlling player       & 19   &  19              \\
		Ball state               & 18   &  18  \\
		Teammates                & 36   &  90           \\
		The closest teammate     & 9    &  9       \\
		Opponents                & 45   &  99   \\
		The closest opponent     & 9    &  9   \\
		Available actions        & 19   &  19    \\
		Match state              & 10   &  10      \\
		Offside judgement        & 10   &  22   \\
		Yellow / Red cards       & 20   &  44   \\
		Sticky action            & 10   &  10     \\
	    Player distance to ball  & 9    &  21     \\
		\bottomrule
	\end{tabular}
	\caption{The content and length of the vector features.}
	\label{table:2}
\end{table}

\section{Details of Network Structure}
\label{appendix:3}
\begin{figure}[h]
	\centering
	\includegraphics[scale=0.22]{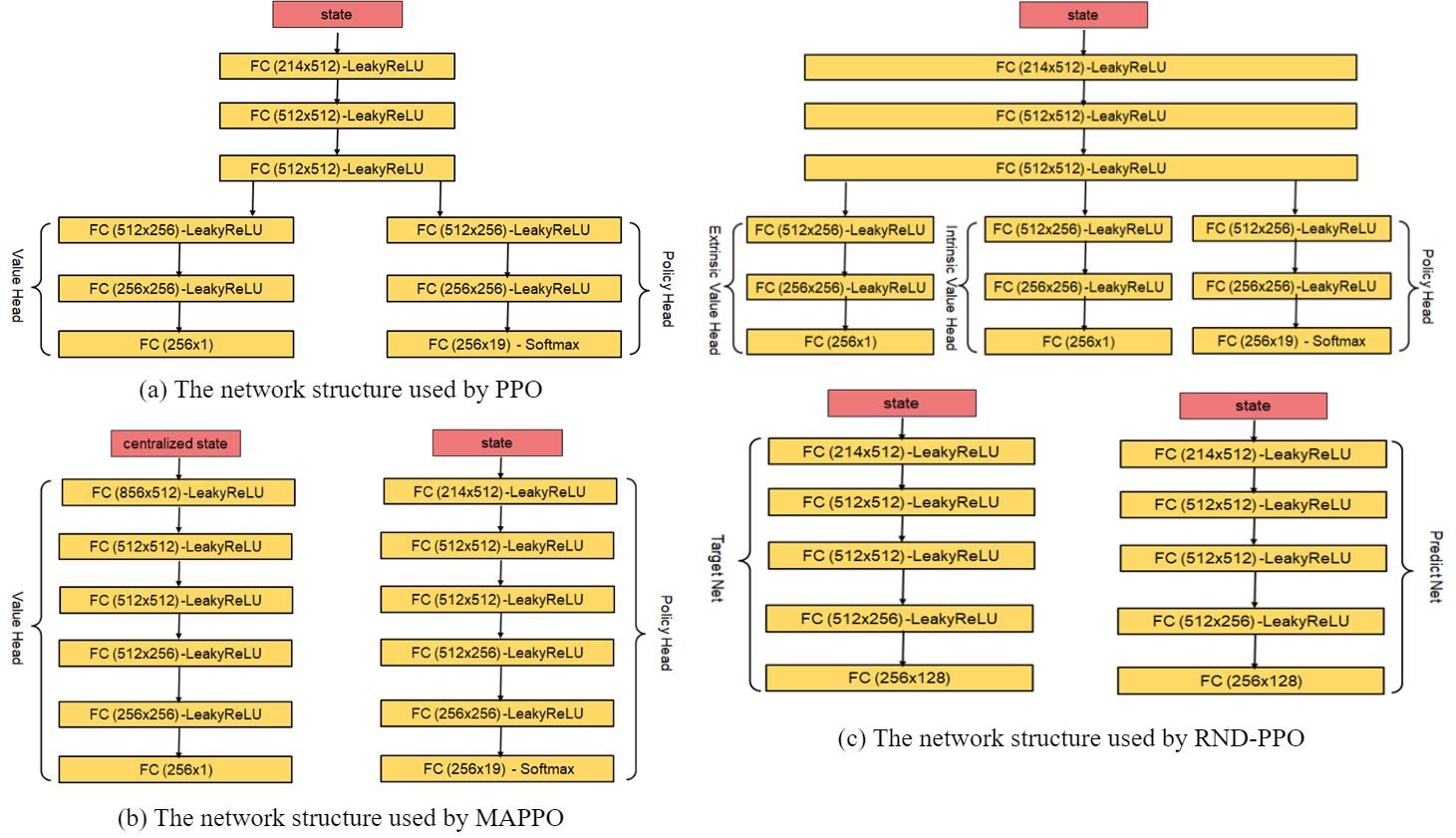}
	\caption{Network structures for the 5v5 training: (a) used by PPO; (b) used by MAPPO; (c) used by RND-PPO.}
	\label{fig:13}
\end{figure}

Figure \ref{fig:13} illustrates the network structures used by PPO, MAPPO, and RND-PPO for the 5v5 training in this paper. For PPO, the policy network and value network comprise six dense layers, and the first three dense layers are shared as the torso network, as depicted in Figure \ref{fig:13}(a). The Softmax layer is utilized to convert the policy head output to the action probability, and the final action is selected by sampling with probability. The activation function Leaky-Relu is applied to the dense layers, except for the last layer. In MAPPO, the policy network and value network are separated, as shown in Figure \ref{fig:13}(b), due to the inconsistency of their input dimensions. The network structure of RND-PPO has three parts, as illustrated in Figure \ref{fig:13}(c). The predict net and target net, and the intrinsic value head, are employed to assess the familiarity of the input state. The network structure for the 11v11 training is similar to that of the 5v5 training, but the network is wider.

\section{Details of the Reward Shaping }
\label{appendix:4}
Table \ref{table:3} presents the specific design details of reward shaping for our proposed method. The team rewards are allocated to all agents in the team, while individual rewards are only granted to the agent who completes the specific task. The main rewards are utilized by all AIs, with other rewards selectively employed by different policy explorers based on their individual training objectives. For instance, one policy explorer incorporates the main rewards and the hold ball reward to train the agent to adopt a more cautious playing style, while another policy explorer employs the secondary attack reward to improve the agent's passing abilities.

\begin{table}[h]
	\centering
	\begin{tabular}{lccc}
		\toprule
		Tasks  &  Value   &  Team   &  Main                          \\
		\midrule
		Goal / Lose goal       & 1 / -1     &  $ \surd $   &  $ \surd $           \\
		Win / Lose             & 2 / -2     &  $ \surd $   &  $ \surd $\\
		Get / Lose possession  & 0.2 / -0.2 &  $ \surd $   &  $ \surd $        \\
		Out of bounds          & -0.001     &            &  $ \surd $    \\
		Hold ball              & 0.0003     &            &   \\
		Secondary attack       & 0.1        &            & \\
		Successful slide       & 0.1        &            &  \\
		\bottomrule
	\end{tabular}
	\caption{Reward shaping of the AIs.}
	\label{table:3}
\end{table}

\section{Details of the Training Hyperparameters}
\label{appendix:5}
Table \ref{table:4} lists the training hyperparameters utilized in our study. These hyperparameters are consistent across all training processes, except for the last three ones, which are exclusively utilized by the RND-PPO algorithm. The entropy coefficient gradually decreases from 0.01 to 0.001 during the main agent's training, while it remains constant at 0.01 during the training of other AIs. It is noteworthy to mention that given the enormous number of possible hyperparameter combinations, we cannot guarantee that the selected hyperparameter values are optimal. However, we can affirm that the AI trained with these hyperparameters can demonstrate superior performance in the GRF environment.

\begin{table}[h]
	\centering
	\begin{tabular}{lcc}
		\toprule
		Hyperparameter           &  Value   &  Algorithm         \\
		\midrule
		Batch size               & 80,000   &  All              \\
		Trajectory length        & 128   &  All  \\
		Sample reuse             & About 1.0   &  All           \\
		PPO clipping             & 0.2    &  All       \\
		PPO dual-clipping        & 3   &  All   \\
		Gradient clipping        & 10   &  All   \\
		Discount factor $ \gamma $& 0.9995   &  All    \\
		GAE discount $ \lambda  $  & 0.95   &  All      \\
		Value loss weight        & 0.5   &  All   \\
		Entropy coefficient      & 0.01  &  All   \\
		Optimizer                &  Adam  &  All     \\
		Learning rate            &  5e-5 &  All     \\
		Adam $ \beta_1, \beta_2 $    &  0.99, 0.999     &  All     \\
		Self-model sampling rate $ \alpha $  & 0.6    &  All     \\
		\makecell[l]{Softmax temperature in sam-\\pling}     & 0.3    &  All     \\
		Capacity of SHMP         & 100    &  All     \\
		\makecell[l]{Discount factor for calculat-\\ing intrinsic advantage} & 0.99    &  RND-PPO     \\
    	\makecell[l]{Coefficient of intrinsic ad-\\vantage}	  & 8    &  RND-PPO     \\
		\makecell[l]{Coefficient of extrinsic ad-\\vantage}   & 2    &  RND-PPO     \\
		\bottomrule
	\end{tabular}
	\caption{The content and length of the vector features.}
	\label{table:4}
\end{table}

\section{Training Process of 11v11 model}
\label{appendix:6}
\begin{figure}[h]
	\centering
	\includegraphics[scale=0.054]{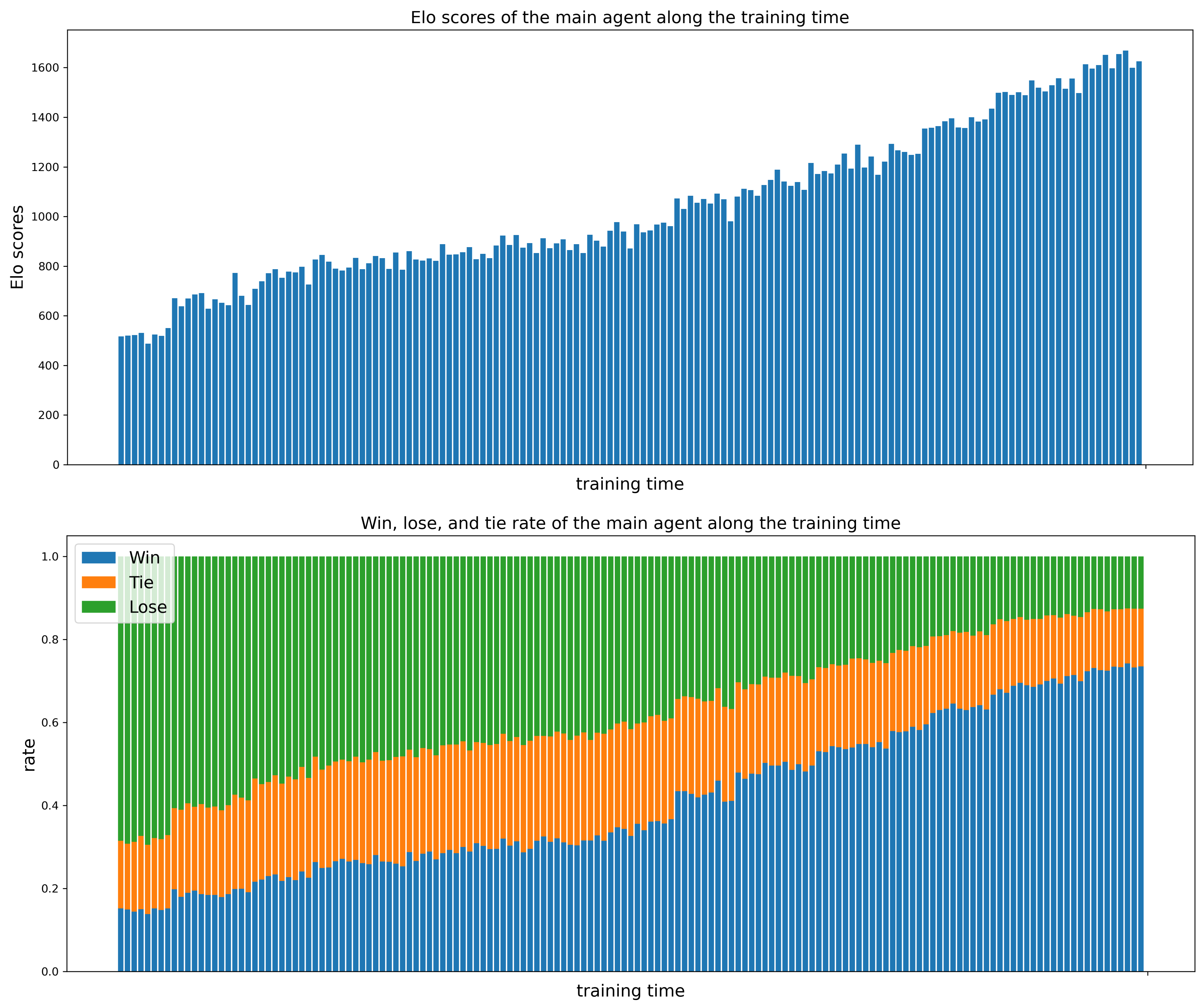}
	\caption{Diagram of the performance change of 11v11 agent with training process, spanning 70 days from the start of training to the end of the competition.}
	\label{fig:14}
\end{figure}

Figure \ref{fig:14} gives the diagram of the performance change of the 11v11 model as the training time increases. The results are similar to the 5v5 training, both the Elo value and win rate of the model have been increasing and the growth trend has not slowed down significantly. The training process performs well in the environments with varying complexity, demonstrating the stability of the proposed DIS.

\section{The Designed Action Mask}
\label{appendix:7}
To avoid the occurrence of meaningless actions during training, such as shooting in one's own penalty area or sliding when no ball is nearby, we designed an action mask in this study. However, we acknowledge that the use of hand-crafted rules in the mask may limit the diversity of strategies employed by the model. Hence, we attempt to restrict as few actions as possible. The action mask comprises the following main components: (1) long pass, high pass, short pass, and shot cannot be used when the player in control is far from the ball; (2) sliding cannot be used when the player is far from the ball and the ball is not under the control of an opposing player; (3) release sprint and release dribble cannot be used when the player is not in a sprint or dribble state, respectively; (4) players are unable to use sprint and dribble actions when they are already in a sprint or dribble state; and (5) the shot action cannot be used in the own half-court. Further details regarding the implementation of the designed action mask can be found in the Supplementary Files.

\end{document}